\documentclass[letterpaper, 10 pt, conference]{ieeeconf}  

\IEEEoverridecommandlockouts                      

\usepackage{epsfig} 
\usepackage{amsmath} 
\usepackage{amssymb}  
\usepackage{color}
\usepackage[linesnumbered,ruled]{algorithm2e}
\usepackage[normalem]{ulem} 
\makeatletter
\let\NAT@parse\undefined
\makeatother
\usepackage{hyperref}
\hypersetup{pdfstartview=FitH,
            colorlinks=true,
            linkcolor=red,
            anchorcolor=blue,
            citecolor=green
            }
\usepackage{cite}
\usepackage{booktabs}
\usepackage{multirow}
\usepackage[caption=false,font=footnotesize]{subfig}
\usepackage[dvipsnames]{xcolor}
\usepackage{subfig}
\usepackage{soul}
\usepackage{balance}
\newcounter{RNum}
\renewcommand{\theRNum}{\arabic{RNum}}
\newcommand{\Remark}{\noindent\textit{\textbf{Remark}~\refstepcounter{RNum}\textbf{\theRNum}: }}
\newcommand{\NoOne}[1]{\textcolor{red}{#1}}
\newcommand{\NoTwo}[1]{\textcolor{green}{#1}}
\newcommand{\NoThree}[1]{\textcolor{blue}{#1}}
\soulregister\NoOne7
\soulregister\NoTwo7
\soulregister\NoThree7
\soulregister\Remark7
\definecolor{table_c}{RGB}{245,228,208}
\title{\LARGE \bf
Enhancing Nighttime UAV Tracking with Light Distribution Suppression
}
\author{Liangliang Yao$^{1}$, Changhong Fu$^{1,*}$, Yiheng Wang$^{1}$, Haobo Zuo$^{2}$, Kunhan Lu$^{1}$ %
\thanks{$^{*}$Corresponding author}
\thanks{
$^{1}$L. Yao, C. Fu, Y. Wang, and K. Lu are with the School of Mechanical Engineering, Tongji University, Shanghai 201804, China.
\itshape{Email: changhongfu@tongji.edu.cn}
}
\thanks{
$^{2}$H. Zuo is with the Department of Computer Science, University of Hong Kong, Hong Kong 999077, China.
}
}
\begin{document}
\maketitle
\thispagestyle{empty}
\pagestyle{empty}
\begin{abstract} Visual object tracking has boosted extensive intelligent applications for unmanned aerial vehicles (UAVs). 
However, the state-of-the-art (SOTA) enhancers for nighttime UAV tracking always neglect the uneven light distribution in low-light images, inevitably leading to excessive enhancement in scenarios with complex illumination. To address these issues, this work proposes a novel enhancer, \textit{i.e.}, LDEnhancer, enhancing nighttime UAV tracking with light distribution suppression. Specifically, a novel image content refinement module is developed to decompose the light distribution information and image content information in the feature space, allowing for the targeted enhancement of the image content information. Then this work designs a new light distribution generation module to capture light distribution effectively. The features with light distribution information and image content information are fed into the different parameter estimation modules, respectively, for the parameter map prediction. Finally, leveraging two parameter maps, an innovative interweave iteration adjustment is proposed for the collaborative pixel-wise adjustment of low-light images. Additionally, a challenging nighttime UAV tracking dataset with uneven light distribution, namely NAT2024-2, is constructed to provide a comprehensive evaluation, which contains 40 challenging sequences with over 74K frames in total. Experimental results on the authoritative UAV benchmarks and the proposed NAT2024-2 demonstrate that LDEnhancer outperforms other SOTA low-light enhancers for nighttime UAV tracking. Furthermore, real-world tests on a typical UAV platform with an NVIDIA Orin NX confirm the practicality and efficiency of LDEnhancer. The code is available at \url{https://github.com/vision4robotics/LDEnhancer}. 

\end{abstract}
\section{Introduction} \label{sec:intro}
Visual object tracking has gained increasing attention in various applications of unmanned aerial vehicle (UAV), \textit{e.g.}, search and rescue missions~\cite{varga2022seadronessee}, obstacle avoidance~\cite{chen2020computationally}, aerial cinematography~\cite{bonatti2019IROS}, and 3D localization~\cite{zhang2019eye}. 
With the position provided in the initial frame, the UAV tracker aims to predict the target location in the following frames. For nighttime UAV tracking, the performance of the trackers~\cite{li2019siamRPN++,cao2021siamapn++,peng2022lpat,cao2021hift,gao2022aiatrack,yao2023sgdvit} designed for high-visibility circumstances presents a significant drop due to the limited illumination, low contrast, and much noise in low-light images~\cite{li2022all}. This issue is particularly pronounced in scenarios with complex illumination. Consequently, developing robust and effective nighttime UAV tracking remains a challenging problem.

Over the past years, the low-light image enhancer based on iteration enhancement~\cite{ye2021darklighter,fu2022highlightnet,ye2022tracker} has been one of the mainstream solutions for nighttime UAV tracking. These methods mainly estimate an enhancement parameter map from the low-light image and adjust low-light images iteratively based on the enhancement map. By boosting the intensity of dark regions, the performance of the UAV trackers trained with daytime images has gained substantial improvement. However, when applied to low-light images captured in scenarios with complex illumination, these methods neglect the uneven light distribution and inevitably further enhance the regions that are already sufficiently bright. The over-enhancement and saturation after low-light image enhancement (LLIE) may destroy the extracted image features, leading to a reduction in image content information which is significant for nighttime UAV tracking. \textbf{\textit{Hence, how to devise a low-light enhancer capable of addressing uneven light distribution is a pressing need for nighttime UAV tracking.}}
\begin{figure}[!t]
\vspace{6pt}
\colorbox{table_c}{
	\centering
	\includegraphics[width=0.98\linewidth]{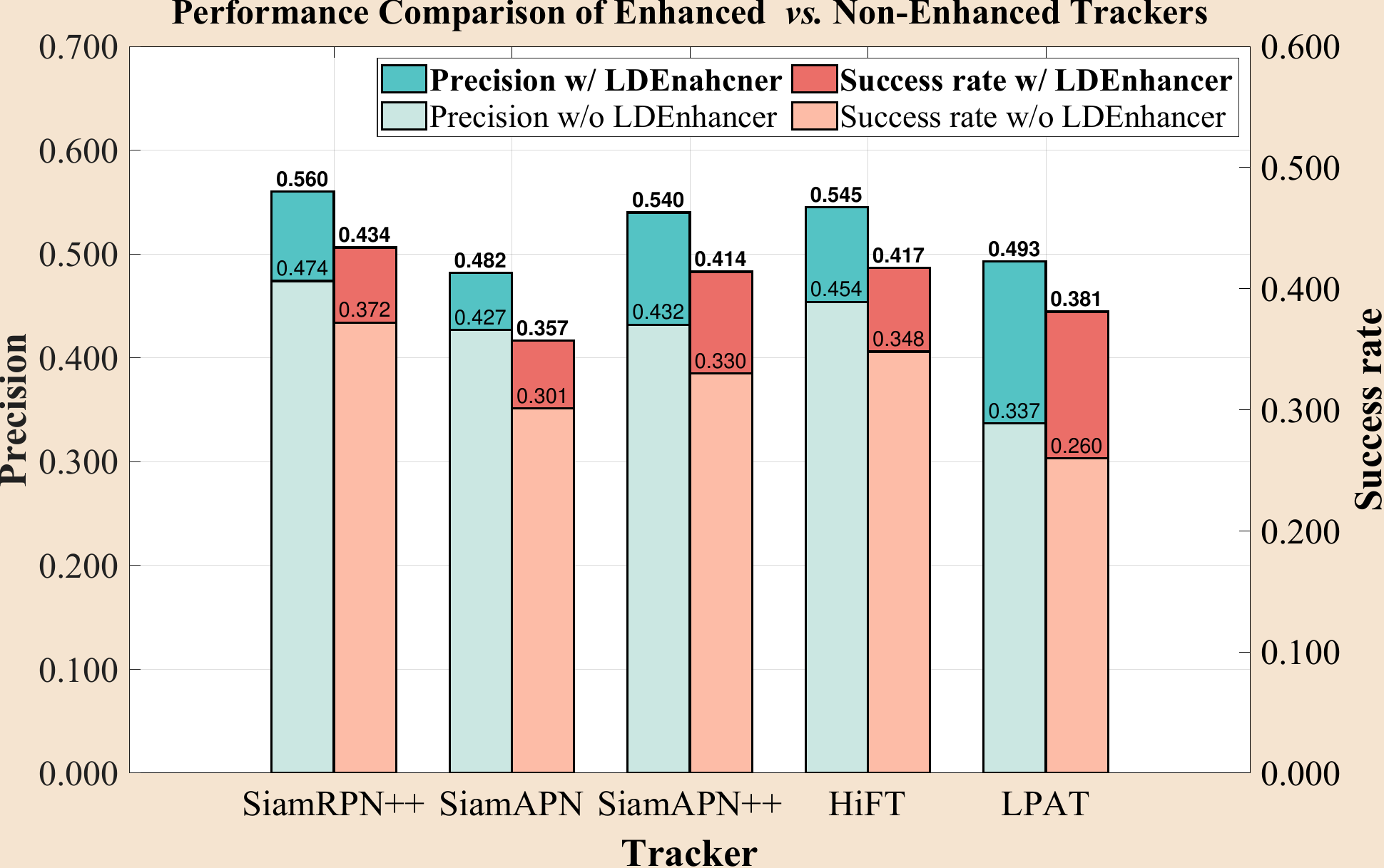}}
	\caption
    {Overall performance of the state-of-the-art (SOTA) trackers~\cite{cao2021siamapn++, cao2021hift, li2019siamRPN++, peng2022lpat, fu2021onboard} with the proposed LDEnhancer enabled or not on UAVDark135~\cite{li2022all}. With the support of LDEnhancer, SOTA UAV trackers, which exhibit difficulty in nighttime scenarios with uneven light distribution, have been significantly improved. SiamAPN++~\cite{cao2021siamapn++} with LDEnhancer gains the improvement with \textbf{25.0\%} and \textbf{25.5\%} in precision and success rate, respectively. Meanwhile, LPAT~\cite{peng2022lpat} gets the promotion with \textbf{46.3\%} and \textbf{46.5\%} in precision and success rate, respectively.
	}
	\label{fig:fig1}
        \vspace{-15pt}
\end{figure}

Recently, light distribution suppression~\cite{li2015nighttime,yan2020nighttime,sharma2021nighttime,jin2022unsupervised} has garnered widespread attention in image processing. The majority of these works~\cite{li2015nighttime,yan2020nighttime} are towards light distribution suppression on haze or foggy nights, posing a challenge to directly utilize in real-world nighttime scenarios. Moreover, they are unable to effectively enhance dark regions in low-light images. Only several approaches~\cite{jin2022unsupervised} have the capability of suppressing bright regions and enhancing dark regions simultaneously. Despite the works generating some visually pleasing results in LLIE, the efficacy for enhancing nighttime UAV tracking remains notably unsatisfactory due to the misalignment of the optimization objectives with nighttime UAV tracking.
\textbf{\textit{Thereby, how to effectively combine a low-light enhancer with light distribution suppression for nighttime UAV tracking is worth exploring carefully.}} 

In this work, a novel enhancer, \textit{i.e.}, LDEnhancer, is proposed to enhance nighttime UAV tracking with light distribution suppression. Figure~\ref{fig:fig1} shows the tracking performance of the state-of-the-art (SOTA) trackers with the proposed LDEnhancer enabled or not on UAVDark135~\cite{li2022all}. With LDEnhancer, the performance of UAV trackers has been greatly improved in low-light conditions. The main contributions are as follows:
\begin{itemize}
	\item A novel low-light image enhancer, dubbed LDEnhancer, is proposed to enhance nighttime UAV tracking with light distribution suppression, where the task is treated as a collaborative pixel-wise adjustment with suppression parameter map and enhancement parameter map.
	\item A novel image content refinement module is designed to decompose the light distribution information and image content information in the feature space, allowing for the targeted enhancement of the image content information. A new light distribution generation module is developed to capture light distribution effectively. 
        \item Two different parameter estimation modules are constructed to process the features with light distribution information and image content information, respectively, for the parameter map prediction. 
	\item Comprehensive evaluation on authoritative nighttime UAV tracking benchmarks and the proposed NAT2024-2 verify the effectiveness and robustness of LDEnhancer for nighttime UAV tracking. Real-world tests demonstrate that LDEnhancer has superior advantages for the deployment of nighttime UAV tracking, especially for scenarios with complex illumination.
\end{itemize}

\section{Related Work}
\subsection{Low-Light Image Enhancement (LLIE)}
The primary objective of LLIE~\cite{guo2020zero, li2021learning, ma2022toward, zhang2021beyond, liu2021retinex,guo2016lime,jiang2021enlightengan,zhang2021learning, xu2023low} is to enhance the perception or interpretability of images captured in low-light environments. C.~Li~\textit{et al.}~\cite{li2021learning} estimate an enhancement parameter map from the input low-light image, which is then used for image enhancement in an iteration manner. KinD++~\cite{zhang2021beyond}, inspired by the Retinex theory, decomposes the images into two parts: illumination map and reflection map. Then this method makes adjustments to the illumination map and removes degradation on the reflection map. R.~Liu~\textit{et al.}~\cite{liu2021retinex} combine Retinex theory and unrolled optimization, using a cooperative prior architecture search algorithm. For nighttime UAV tracking, several approaches~\cite{ye2021darklighter,ye2022tracker,fu2022highlightnet} have been proposed to combine LLIE with UAV trackers for enhancing performance. Darklighter~\cite{ye2021darklighter} designs a lightweight convolutional neural network (CNN)-based enhancement parameter map estimation network to cope with poor illumination and intricate noise. HighlightNet~\cite{fu2022highlightnet} proposes an iteration-based low-light enhancer to facilitate both the online target selection step and tracking step for nighttime UAV tracking. J.~Ye~\textit{et al.}~\cite{ye2022tracker} construct a low-light enhancer with a spatial-channel Transformer to achieve favorable LLIE iteratively. However, the SOTA enhancers tend to significantly overenhance bright regions in captured images under conditions of complex illumination, resulting in a suboptimal performance for nighttime UAV tracking.

\subsection{Light Distribution Suppression}
Light distribution suppression~\cite{li2015nighttime,yan2020nighttime,sharma2021nighttime,jin2022unsupervised}, a challenging problem to be resolved in image processing, has caught worldwide research attention.  
In night image dehazing, a few methods have been proposed to suppress light distribution. 
Y.~Li~\textit{et al.}~\cite{li2015nighttime} model the glow component by convolving the light distribution with the atmosphere point spread function, and decomposing it from the image.
J.~Zhang~\textit{et al.}~\cite{zhang2017fast} refine the hazy
imaging model with uneven light distribution, which is estimated by maximum reflectance prior. 
W.~Yan~\textit{et al.}~\cite{yan2020nighttime} defog the high and low frequency components of grayscale images and guide light distribution suppression of colored images with defogged grayscale images.
A.~Sharma~\textit{et al.}~\cite{sharma2021nighttime} propose a novel approach grounded in camera response function estimation and high dynamic range imaging to mitigate light distribution. For LLIE with light distribution suppression, Y. Jin~\textit{et al.}~\cite{jin2022unsupervised} propose an unsupervised night image enhancement method that combines layer decomposition and light-effects suppression to achieve better visibility, color fidelity, and naturalness in low-light conditions without manual annotations or prior knowledge. However, most works in night image dehazing are only designed for light distribution suppression, without considering enhancing dark regions simultaneously. Moreover, the misalignment of the optimization objectives between LLIE and nighttime UAV tracking poses a limitation to enhancing the tracking performance in low-light scenarios.

\begin{figure*}[!t]	
	\centering
	\includegraphics[width=1\linewidth]{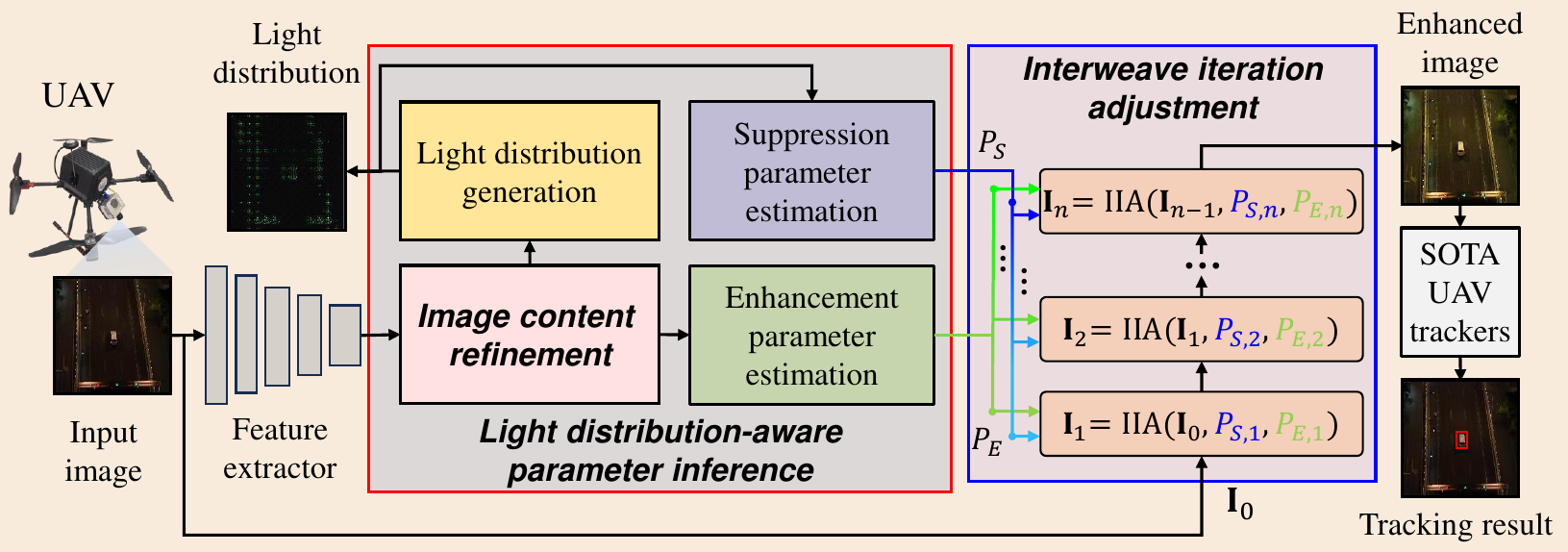}
        \setlength{\abovecaptionskip}{-4pt} 
        \vspace{-0.3cm}
	\caption
        {Overview of our proposed LDEnhancer. The LDEnhancer is composed of three parts, from left to right: \emph{feature extraction}, \emph{light distribution-aware parameter inference}, and \emph{interweave iteration enhancement.} $P_S$ denotes the suppression parameter map. $P_E$ represents the enhancement parameter map. }
	\label{fig:main}
        \vspace{-10pt}
\end{figure*}

\begin{table}[!b]
    \centering
    \caption{Details of the feature extractor.}
    \setlength{\tabcolsep}{0.5mm}
    \resizebox{\linewidth}{!}{
    \colorbox{table_c}{
    \begin{tabular}{ccccccc}
         \toprule
         Input & Data dimensions & Operator
         & Kernel & Stride&Padding & Output \\
         \midrule
        $\mathbf{I}_0$ & $256\times256\times3$ & CNR & $2\times2$ &2&0&$\mathbf{C}_1$\\
         $\mathbf{C}_1$ & $128\times128\times4$ & CNR & $2\times2$ &2&0&$\mathbf{C}_2$\\
         $\mathbf{C}_2$ & $64\times64\times4$ & CNR & $2\times2$ &2&0&$\mathbf{C}_3$\\
         $\mathbf{C}_3$ & $32\times32\times8$ & CNR & $2\times2$ &2&0&$\mathbf{F}_0$\\
          \bottomrule
    \end{tabular}
    }
    }
    \label{tab:1}%
\end{table}

\section{Proposed Method}
In this section, the proposed enhancer, LDEnhancer, is introduced in detail. As depicted in Fig.~\ref{fig:main}, the proposed framework can be partitioned into three parts, \textit{i.e.}, \emph{feature extraction}, \emph{light distribution-aware parameter inference}, and \emph{interweave iteration adjustment.}

\subsection{Feature Extraction}
During feature extraction, this work constructs a CNN-based feature extractor to extract robust features from input images. TABLE~\ref{tab:1} shows the detailed network settings of the feature extractor. As an illustrative example, an input with data dimensions of $256\times256\times3$ is considered. The network architecture comprises four lightweight CNR layers, where CNR denotes the sequential operations of a convolutional layer, batch normalization, and a rectified linear unit (ReLU) activation function. To extract high-level features while reducing the size of the feature maps, the feature extractor employs a convolutional layer with a $2\times2$ kernel and a stride size of 2 as a down-sampling operation.  

\subsection{Light Distribution-Aware Parameter Inference}
In light distribution-aware parameter inference, this work designs a novel image content refinement module to process the feature generated from the feature extractor. Then the decomposed features are posted to a new light distribution generation module. The light distribution generation module captures light distribution effectively with a proposed unsupervised light loss. The resulting features containing image content information and light distribution information are then used in separate estimation modules for suppression and enhancement parameter map prediction.

\begin{figure}[!t]	
    \vspace{5pt}
    \centering
    \includegraphics[width=1\linewidth]{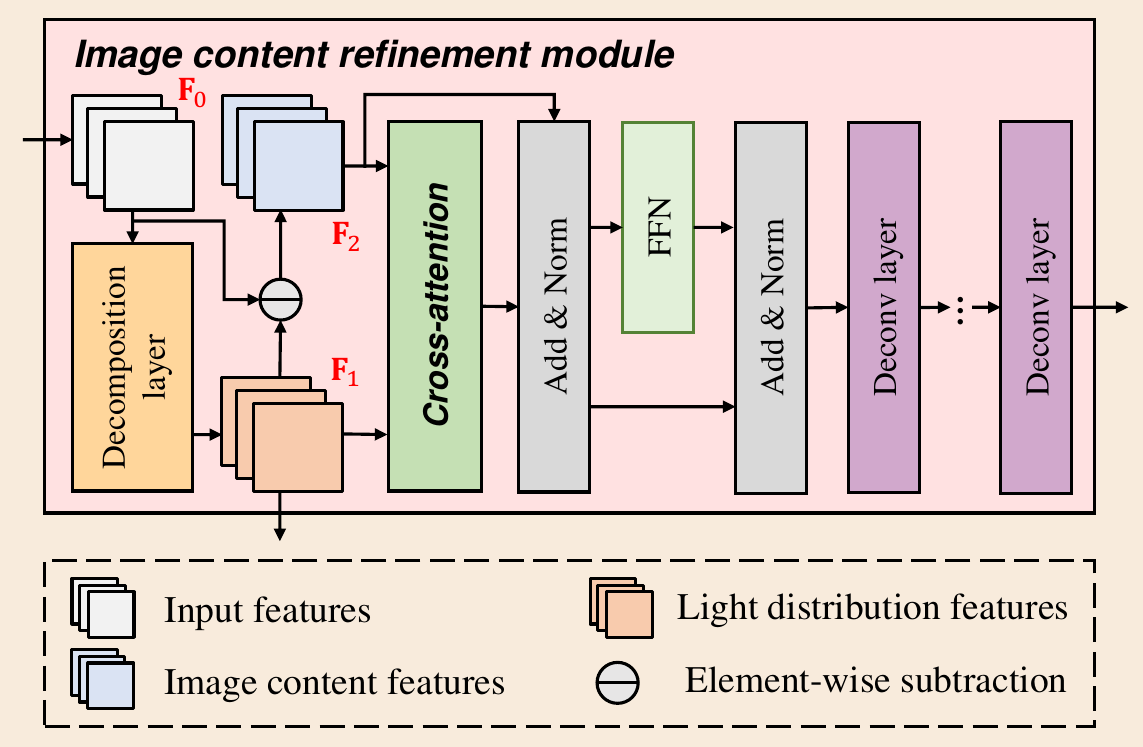}
    \setlength{\abovecaptionskip}{-4pt} 
    \caption
    {
    Detailed workflow of the image content refinement module. 
    }
    \label{fig:iros2024_3}
    \vspace{-15pt}
\end{figure}

\noindent\textbf{Image content refinement}.
As shown in Fig.~\ref{fig:iros2024_3}, the image content refinement module takes the features ${\mathbf{F}_0}$ generated from the feature extractor as the input. The features ${\mathbf{F}_0}$ are fed into the proposed decomposition layer $\mathrm{Decomposition}$ for generating light distribution features ${\mathbf{F}_1}$. Specifically, the decomposition layer is a $1\times1$ convolution layer. Then subtract light distribution features ${\mathbf{F}_1}$ from the input features ${\mathbf{F}_0}$ element by element. The results are denoted as ${\mathbf{F}_2}$. The whole process is defined as:

\begin{equation}\label{1}
    \begin{split}
    \mathbf{F}_1 &= \mathrm{Decomposition}(\mathbf{F}_0)\quad\quad,\\
    \mathbf{F}_2 &=\mathbf{F}_0 - \mathbf{F}_1\quad\quad.
    \end{split}
\end{equation}

The cross-attention $\mathrm{CA}$ uses the feature $\mathbf{F}_2$ extracted by the feature extraction network as $\mathbf{Q}$ (Query), while the estimated light distribution $\mathbf{F}_1$ as $\mathbf{K}$ (Key) and $\mathbf{V}$ (Value). To clarify, the position encoding is denoted as $\mathbf{P}$. The results of the cross-attention are fed into the deconvolution layers $\mathrm{DeConv}$ to adjust the dimension. The output $\mathbf{O}_1$ of the image content refinement module is obtained by:
\begin{equation}\label{2}
	\begin{split}
	\mathbf{M}_{\rm x} &= \mathrm{CA}(\mathbf{F}_2+\mathbf{P},\mathbf{F}_1+\mathbf{P},\mathbf{F}_1) \quad\quad,\\
	\mathbf{M}_{\rm y}&=\mathrm{Norm}(\mathbf{M}_{\rm x}+\mathbf{F}_2) \quad\quad, \\
        \mathbf{O}_1&= \mathrm{DeConv}(\mathrm{Norm}(\mathrm{FFN}(\mathbf{M}_{\rm y})+\mathbf{M}_{\rm y}))\quad\quad,
	\end{split}
\end{equation}
where $\mathbf{M}_{\rm x}$ and $\mathbf{M}_{\rm y}$ are the intermediate variable. $\mathrm{CA}$ represents the procedure of the multi-head attention~\cite{dosovitskiy2020vit}. $\mathrm{Norm}$ is the normalization operation. Besides, $\mathrm{FFN}$ stands for feed-forward network. Generally, the cross-attention $\mathrm{CA}$ can be calculated as follows:

\begin{equation}\label{3}
		\begin{split}
            \mathrm{Att}(\mathbf{Q},\mathbf{K},\mathbf{V})&= \mathrm{Softmax}(\frac{\mathbf{Q}\mathbf{K}^\top}{\sqrt{c}})\mathbf{V} \quad\quad,\\
			\mathrm{CA}(\mathbf{Q},\mathbf{K},\mathbf{V}) &= (\mathrm{Cat}(a^1,a^2,...,a^n))\mathbf{W}_{\rm c} \quad\quad, \\
			a^j &= \mathrm{Att}(\mathbf{Q}\mathbf{W}_1^j,\mathbf{K}\mathbf{W}_2^j,\mathbf{V}\mathbf{W}_3^j) \quad\quad,
   \end{split}
\end{equation}
where $\mathrm{Att}$ represents the scaled dot-product attention and $\sqrt{c}$ is the scaling factor to avoid gradient vanishment in the softmax function. $\mathbf{W}_{\rm c} \in\mathbb{R}^{C \times C} $, $\mathbf{W}_{\rm 1}^j \in\mathbb{R}^{C \times C_{M}}$,$\mathbf{W}_{\rm 2}^j \in\mathbb{R}^{C \times C_{M}}$,$\mathbf{W}_{\rm 3}^j \in\mathbb{R}^{C \times C_{M}}$ can all be regarded as fully connected layer operation, where $C_{M}=C/M$, and $M$ is the number of parallel attention head. 

\Remark Utilizing $1\times1$ convolution facilitates the efficient mapping of extracted features to a new feature space, thereby improving the subsequent adaptive generation of light distribution information. Under the guidance of the estimated light distribution features, the image content refinement module is capable of adaptively refining the features containing image content information by attention mechanism. 

\noindent\textbf{Light distribution generation}.
The light distribution generation module is utilized to capture light distribution based on the illumination-prior knowledge. TABLE~\ref{tab:2} shows the details of the light distribution generation module. The module is composed of four lightweight DCNR layers. Specifically, DCNR represents the combination operations of the deconvolution layer, batch normalization, and ReLU.

To address the issue of ambiguity output by the light distribution generation module, it is crucial to provide an appropriate initial estimation. To this end, a light distribution label denoted as $\mathbf{I}_l$ is calculated using illumination-prior knowledge with the relative smoothness method~\cite{li2014single}. Since light distribution is usually smooth variations, this method extracts the light distribution from the input image by applying a second-order Laplacian filter. The unsupervised light distribution loss $\mathcal{L}_l$ to constrain the output $\mathbf{O}_2$ of the light distribution generation module is defined as:
\begin{equation}\label{eq:1}
    \mathcal{L}_l = \mathrm{SmoothL1}(\mathbf{O}_2,\mathbf{I}_l)\quad\quad.
\end{equation}

\Remark The feature extractor and light distribution generation module are combined to form a U-Net-like network architecture, serving as the encoder and decoder, respectively. The feature extractor encodes the input image to extract robust features, while the light distribution generation module reconstructs these features to capture the light distribution.

\begin{table}[!b]
    \centering
    
    \caption{Details of the light distribution generation module.}
    \setlength{\tabcolsep}{0.5mm}
    \resizebox{\linewidth}{!}{
    \colorbox{table_c}{
    \begin{tabular}{ccccccc}
     \toprule
     Input & Data dimensions & Operator
     & Kernel & Stride&Padding & Output \\
     \midrule
    $\mathbf{F}_1$ & $16\times16\times8$& DCNR & $2\times2$ &2&0&$\mathbf{D}_1$\\
     $\mathbf{D}_1$ & $32\times32\times8$ & DCNR & $2\times2$ &2&0&$\mathbf{D}_2$\\
     $\mathbf{D}_2$ & $64\times64\times4$ & DCNR & $2\times2$ &2&0&$\mathbf{D}_3$\\
     $\mathbf{D}_3$ & $128\times128\times4$ & DCNR & $2\times2$ &2&0&$\mathbf{O}_2$\\
     \bottomrule
    \end{tabular}
    }
    }
    \label{tab:2}%
\end{table}


	 


\begin{table}[!b]
    \centering
    \caption{Details of the parameter estimation modules.}
	\setlength{\tabcolsep}{0.5mm}
	\resizebox{\linewidth}{!}{
 \colorbox{table_c}{
    \begin{tabular}{ccccccc}
         \toprule
         Input & Data dimensions & Operator
         & Kernel & Stride&Padding & Output \\
         \midrule
         $\mathbf{O}_2$ & $256\times256\times3$ & CR & $3\times3$ &1&1&Conv1\\
        Conv1 & $256\times256\times16$ & CR& $3\times3$ &1&1&Conv2\\
        Conv2 & $256\times256\times16$ & CR & $3\times3$ &1&1&Conv3\\ 
        Conv3$\&$Conv2& $256\times256\times32$ & CR & $3\times3$ &1&1&Conv4\\
        Conv4$\&$Conv1& $256\times256\times32$ & CT & $3\times3$ &1&1&$P_S$\\
    
         \bottomrule
    \end{tabular}
    }}
    \label{tab:3}%
\end{table}

\noindent\textbf{Parameter estimation}.
The parameter estimation modules are constructed to predict the parameter maps. In this work, the suppression parameter estimation module and enhancement parameter estimation module share a common structure. As shown in TABLE~\ref{tab:3}, the parameter estimation modules consist of five layers. In the first four layers, each CR layer denotes convolution layers with kernel size 3$\times$3 and stride 1, followed by the ReLU. The last CT layer represents convolution layers with kernel size 3$\times$3 and stride 1, followed by the Tanh activation function. To clarify, take the initially captured light distribution $\mathbf{O}_2$ as an instance, the suppression parameter estimation module generates the suppression parameter map. 

\subsection{Interweave Iteration Adjustment}
The interweave iteration adjustment (IIA) is proposed for the collaborative pixel-wise adjustment of low-light images, leveraging two parameter maps.
To avoid information loss due to overexposure in the bright region, the formula is first defined with two parameter maps:
\begin{equation}\label{5}
	\mathbf{I}_n = \mathrm{IIA}(\mathbf{I}_{n-1}, P_{S,n}, P_{E,n})\quad\quad.
\end{equation}

By taking the light distribution into account, the above formula is then revised to: 
\begin{equation}\label{6}
\begin{split}
    \mathbf{I}_n = \mathbf{I}_{n-1} + (P_{E,n}-P_{S,n})\mathbf{I}_{n-1}(1-\mathbf{I}_{n-1})\quad\quad,
\end{split}
\end{equation}
where $\mathbf{I}_n$ represents the result for iteration $n$.

\Remark In a collaborative way, each iteration process suppresses the bright areas and enhances the darker areas in the image simultaneously. The interweave iteration adjustment facilitates convergence toward the desired enhancement results over successive iterations.
\subsection{Loss Functions}
This work adopts five loss functions to train the framework.
Specifically, this work adopts the three same loss functions as DCE++~\cite{li2021learning}, \textit{i.e.}, spatial consistency loss $\mathcal{L}_{\rm spa}$, color constancy loss $\mathcal{L}_{\rm col}$, illumination smoothness loss $\mathcal{L}_{\rm tv}$. For spatial consistency loss, this loss encourages spatial coherence of the enhanced image by preserving the difference of neighboring regions. Considering the impact of light distribution information on image enhancement, this work subtracts light information labels $\mathbf{I}_{l}$ from the original images, using these labels $\mathbf{I}_{o}$ in place of the original images. For illumination smoothness loss, the work adds this loss to each parameter map for interweave iteration adjustment, \textit{i.e.}, $P_{E,n}-P_{S,n}$. Meanwhile, this work proposes a novel image exposure control loss $\mathcal{L}_{\rm ie}$. The loss measures the distance between the average intensity value of a local
region to level K. The $\mathcal{L}_{\rm ie}$ can be expressed as:
\begin{equation}\label{6}
\begin{split}
    \mathcal{L}_{\rm ie} = (\sum_{i=1}^{T}\mathrm{SmoothL1}(\alpha\mathrm{E}_i,\mathrm{K}))/T\quad\quad,
\end{split}
\end{equation}
where $\alpha$ stands for the coefficient, $T$ represents the number of nonoverlapping local regions of size 16×16, and $\mathrm{E}_i$ is the average intensity value of a local region in the enhanced image. To sum up, the overall loss function of LDEnhancer is presented as:
 \begin{equation}\label{equ:overall} 
	\begin{split}
		\mathcal{L}_{\rm all} =\lambda_1 \mathcal{L}_{\rm spa} + \lambda_2 \mathcal{L}_{\rm col} + \lambda_3 \mathcal{L}_{\rm tv} + \lambda_4 \mathcal{L}_{\rm ie}  + \lambda_5 \mathcal{L}_{l}
	\end{split}
	\quad,
\end{equation}
where $\lambda_1, \lambda_2, \lambda_3, \lambda_4, \lambda_5$ are the coefficients of the loss function. In this work, the value of the coefficients adopts 10, 5, 1, 10, 1, respectively.

\Remark Substantial low-light images encompassing uneven light distribution are an obstacle to training an enhancer with fully supervised learning-based methods. Hence, this work proposes an unsupervised loss to train the network. 

\section{NAT2024-2 Benchmark} 
For comprehensive evaluation in real nighttime scenarios, a new nighttime UAV tracking benchmark with uneven Light distribution, \textit{i.e.}, NAT2024-2, is constructed. It consists of 40 long image sequences, which own 74K frames in total. Each sequence is over \textbf{1500} frames. The sequence categories in the NAT2024-2 dataset encompass diverse targets (\textit{e.g.}, cars, trucks, persons, buses, subways, traffic signs, buildings, and ships) or activities (\textit{e.g.}, cycling, running, and ball games). Employing long-term nighttime UAV tracking benchmarks can better emulate the intricate real-world scenarios, aiding in assessing the performance in complex environments.
\begin{figure}[!t]
    \centering
    \vspace{5pt}
    \includegraphics[width=1\linewidth]{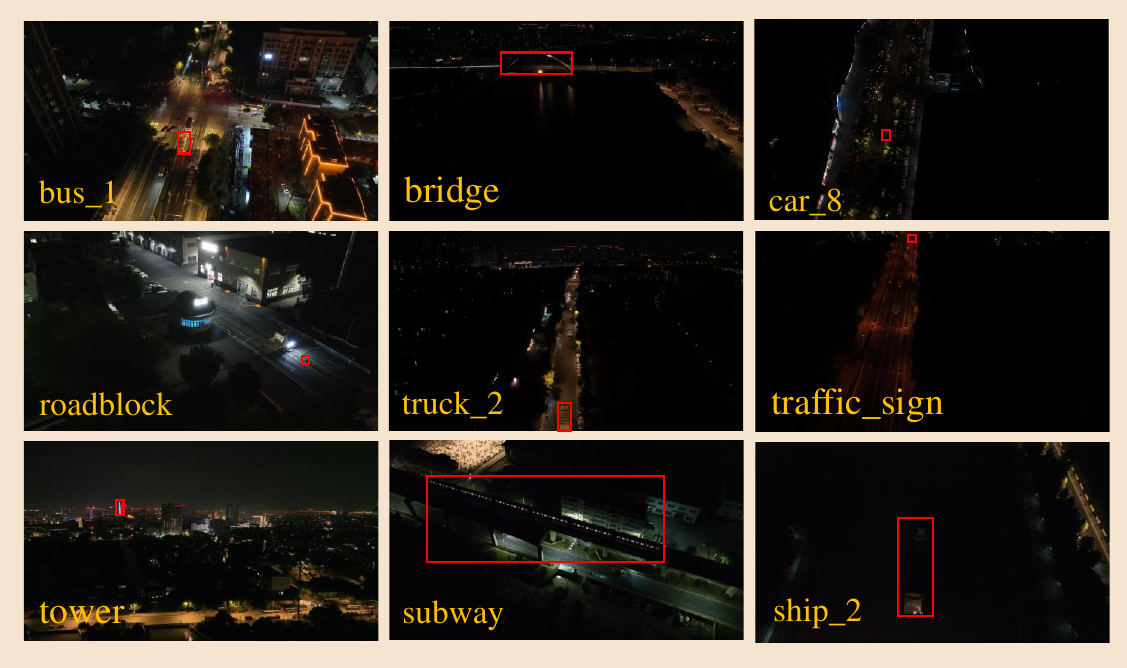}
    \vspace{-20pt}
    \caption{First frames of sequences from NAT2024-2. The tracking object is marked with \textcolor{red}{red} boxes.}
    \label{fig:dataset}
    \vspace{-10pt}
\end{figure}
\begin{figure*}[!t]	

\colorbox{table_c}{
    \subfloat[Precision, normalized precision, and success plots on UAVDark135.]{
	\includegraphics[width=0.32\linewidth]{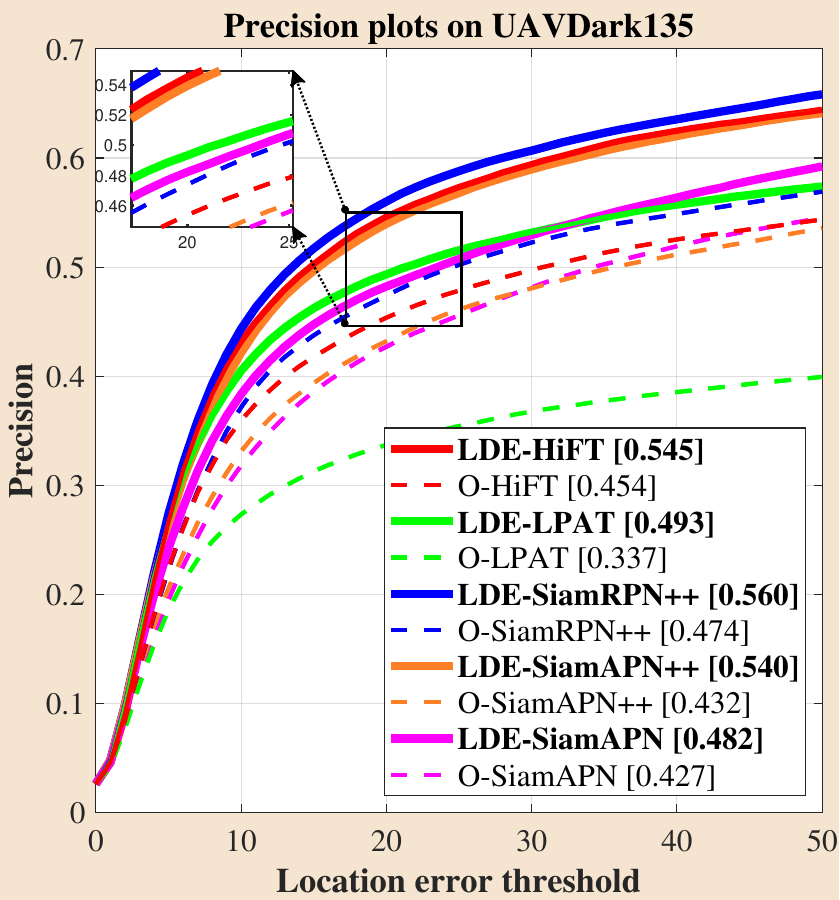}
	\includegraphics[width=0.32\linewidth]{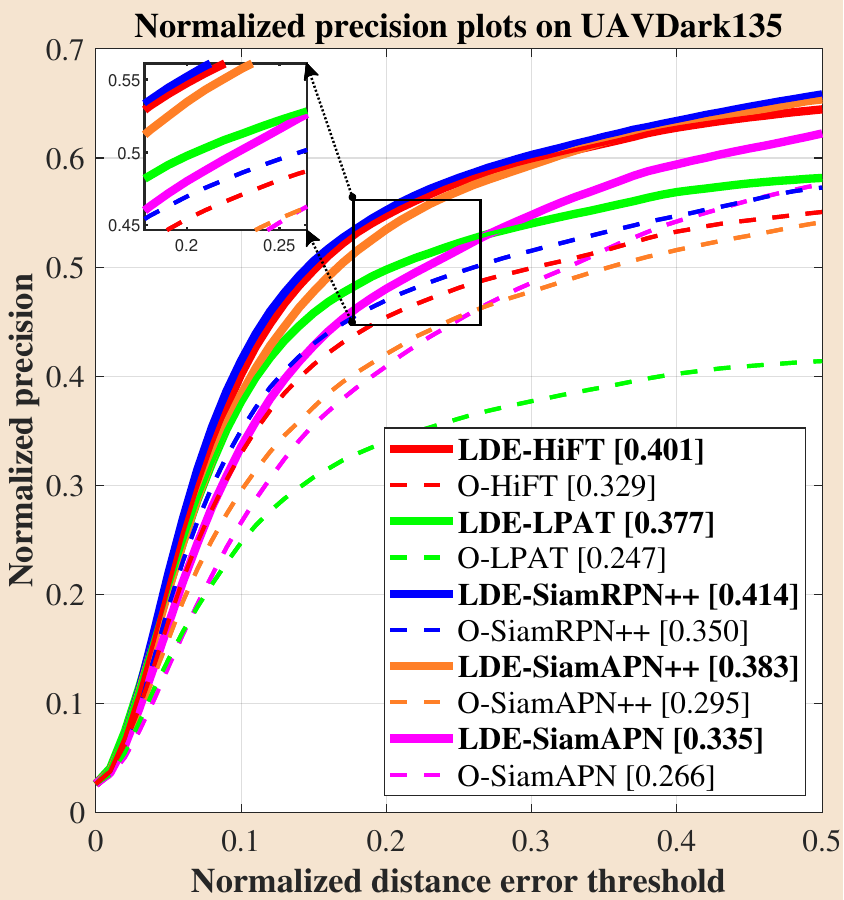}
	\includegraphics[width=0.32\linewidth]{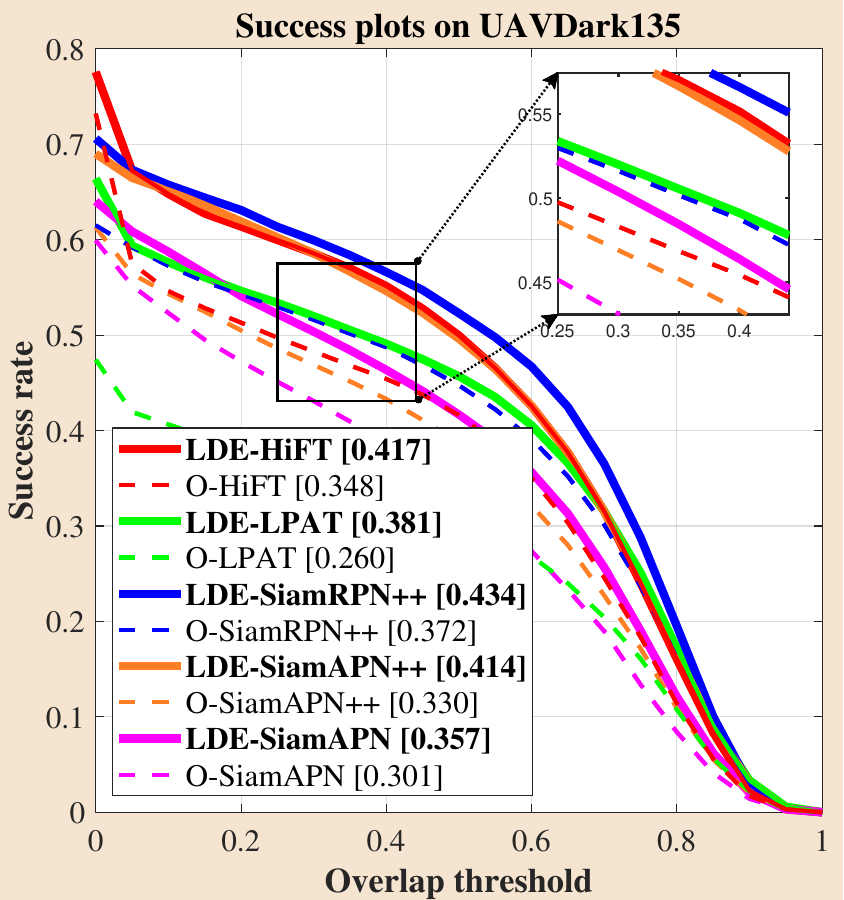}}}
   \colorbox{table_c}{
    \subfloat[Precision, normalized precision, and success plots on NAT2024-2.]{
    \includegraphics[width=0.32\linewidth]{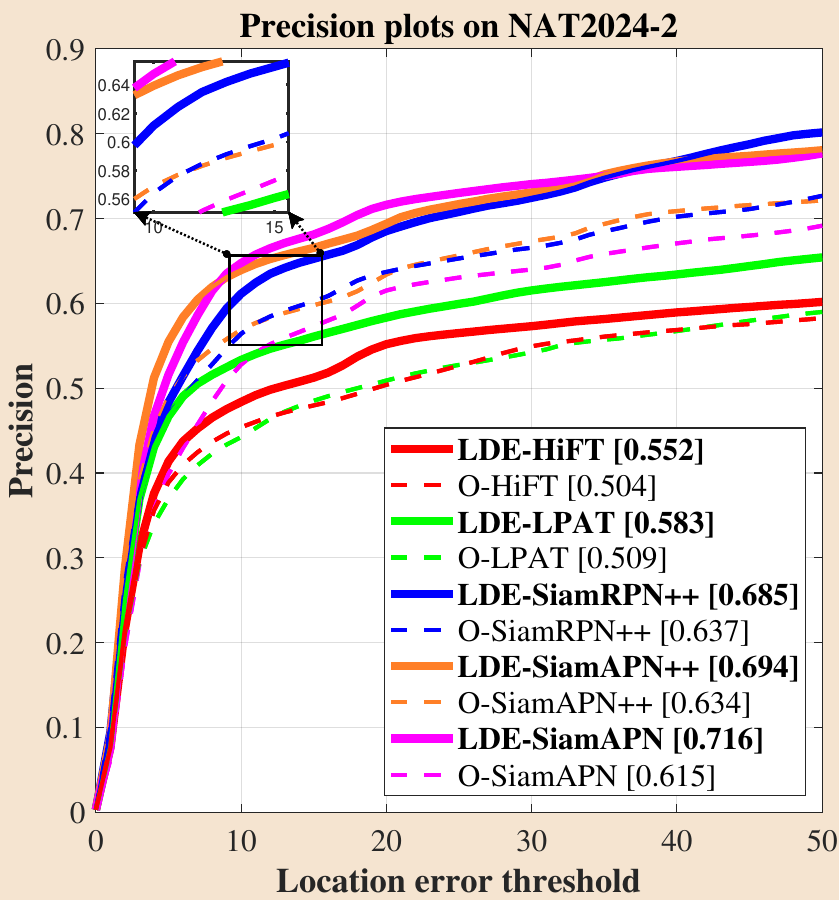}
	\includegraphics[width=0.32\linewidth]{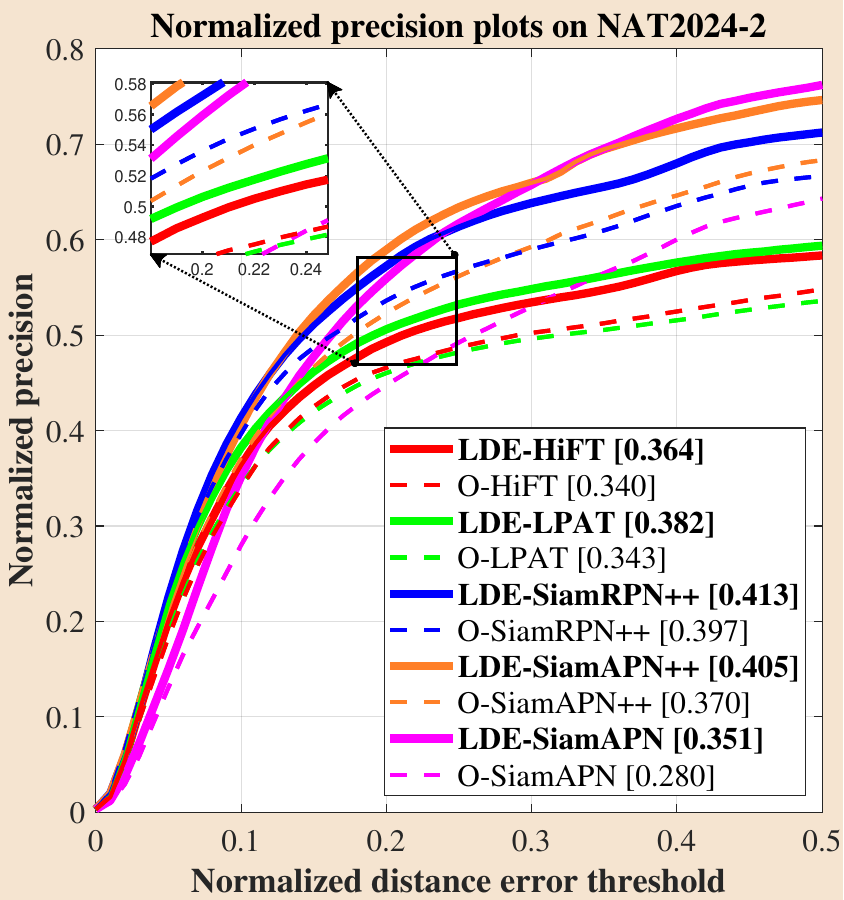}
	\includegraphics[width=0.32\linewidth]{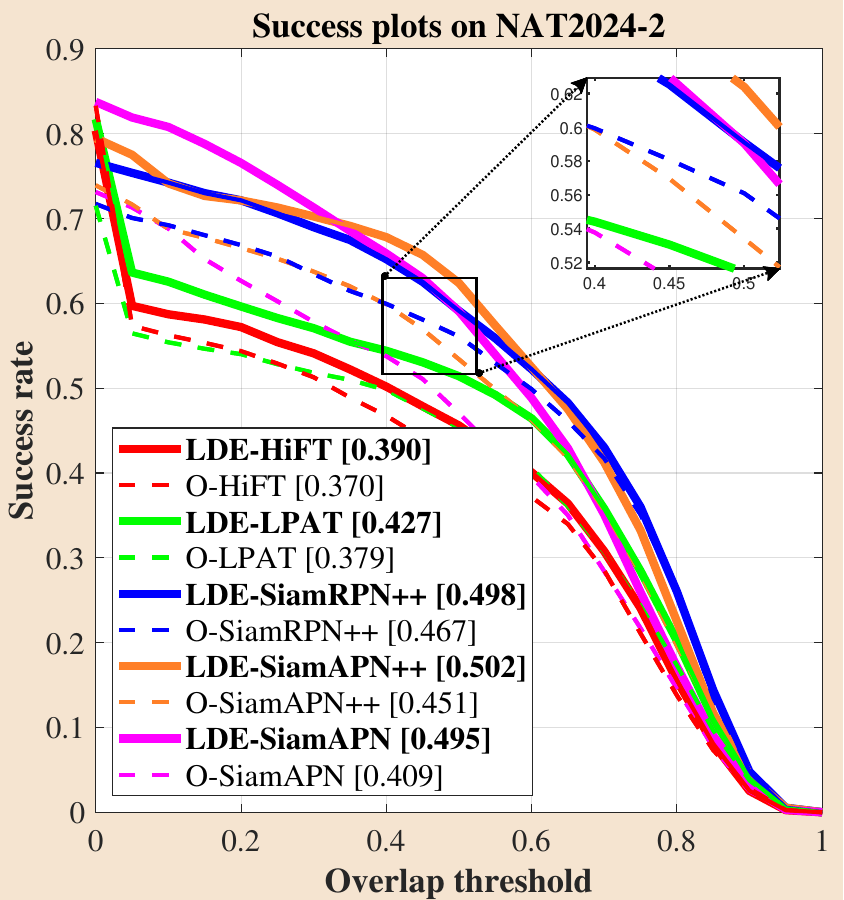}}}
	\caption
	{
		 Overall performance of SOTA UAV trackers with LDEhancer utilized (plots with solid lines) or not (plots with dashed lines). The same color denotes the same tracker. The results demonstrate the effectiveness and superiority of the proposed methods for nighttime UAV tracking. 
	}
	\label{fig:all}
	\vspace{-12pt}
\end{figure*}

\begin{figure}[!t]
        \vspace{6pt}
	\centering
	\includegraphics[width=1\linewidth]{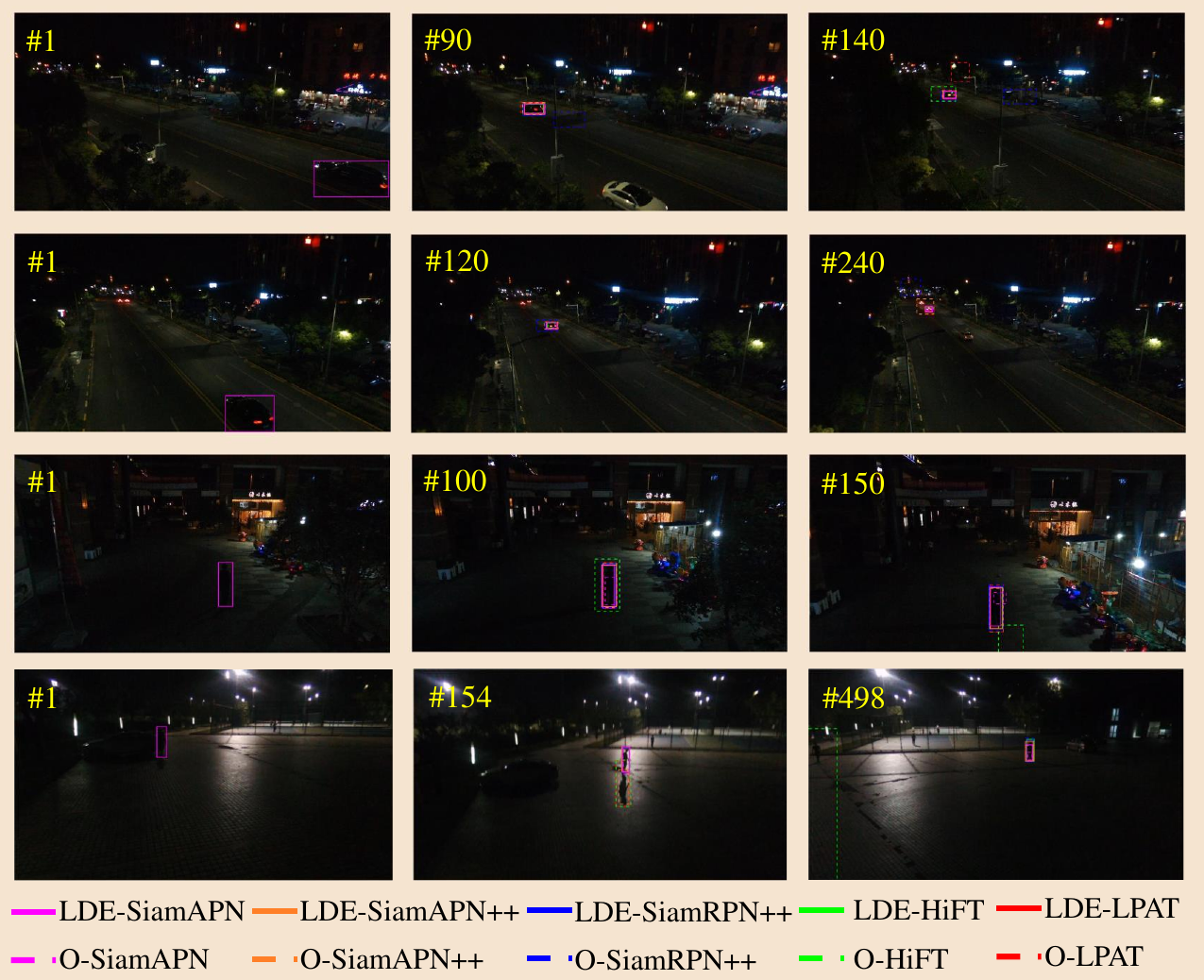}
         \vspace{-0.7cm}
	\caption
        {
        Visualization of trackers with LDEnhancer activated (solid boxes with the prefix “LDE”) or not (dashed boxes with the prefix "O"). From up to down, the sequences are \textit{car4}, \textit{car6}, \textit{pedestrian5\_3}, and \textit{person8} from the UAVDark135~\cite{li2022all}. With LDEnhancer, the trackers achieve robust tracking.
        }
        \label{fig:line}
        \vspace{-0.8cm}
\end{figure}

\section{Experiments}
In this section, the overall performances of the LDEnhancer in nighttime UAV tracking are presented, demonstrating the effectiveness and superiority of LDEnhancer. Real-world tests are conducted to validate the practical feasibility. 

\subsection{Implementation Details}
The proposed LDEnhancer is trained for 100 epochs with PyTorch on a single NVIDIA A100 GPU. For training data, the random sampling strategy is utilized to process the real nighttime UAV tracking dataset. The images are sampled from the train set of NAT2021-$\it{train}$~\cite{ye2022unsupervised} every 50 frames. The training dataset is denoted as NAT2021-$\it{light}$. In the training process, images are first resized to 256 × 256. The AdamW optimizer~\cite{loshchilov2017decoupled} with a fixed learning rate of 0.000001 is adopted to optimize model parameters with a weight decay of 0.0001. The batch size is set to 4. 

\begin{figure*}[!t]
	\centering
 \colorbox{table_c}{
	\begin{minipage}{0.235\linewidth}
		\centering
		\includegraphics[width=1.0\linewidth]{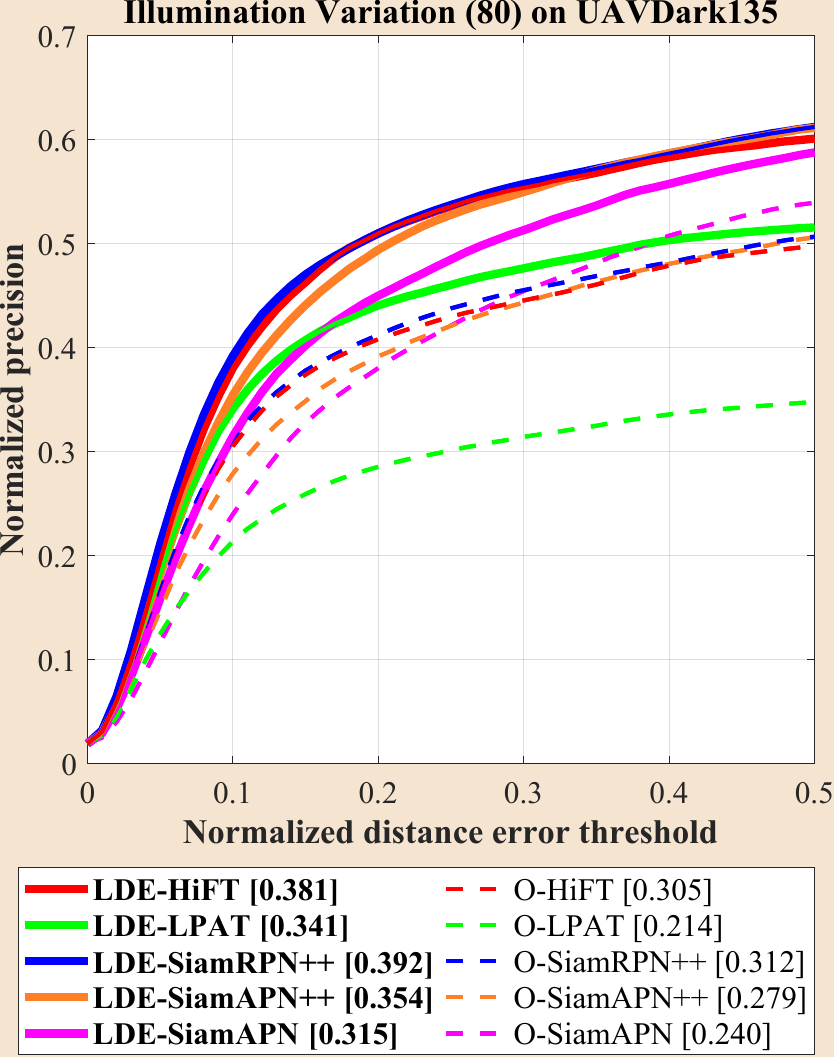}
	\end{minipage}
	\begin{minipage}{0.235\linewidth}
		\centering
		\includegraphics[width=1.0\linewidth]{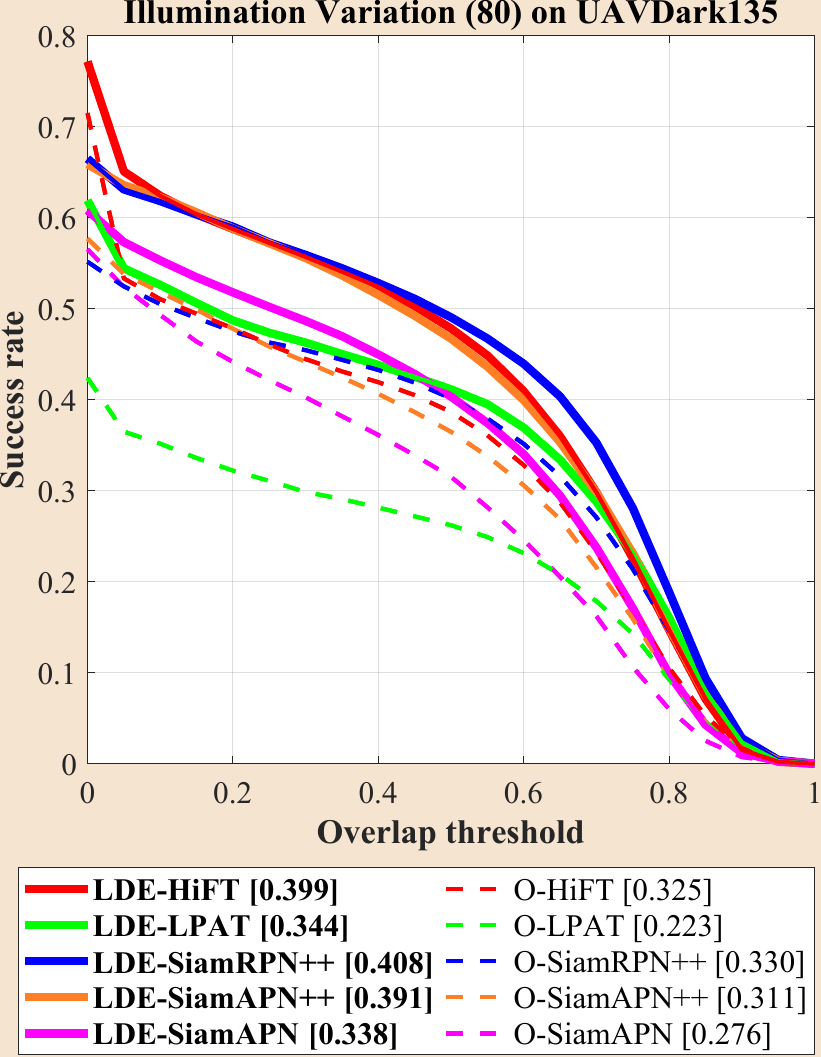}
	\end{minipage}
	\begin{minipage}{0.235\linewidth}
		\centering
		\includegraphics[width=1.0\linewidth]{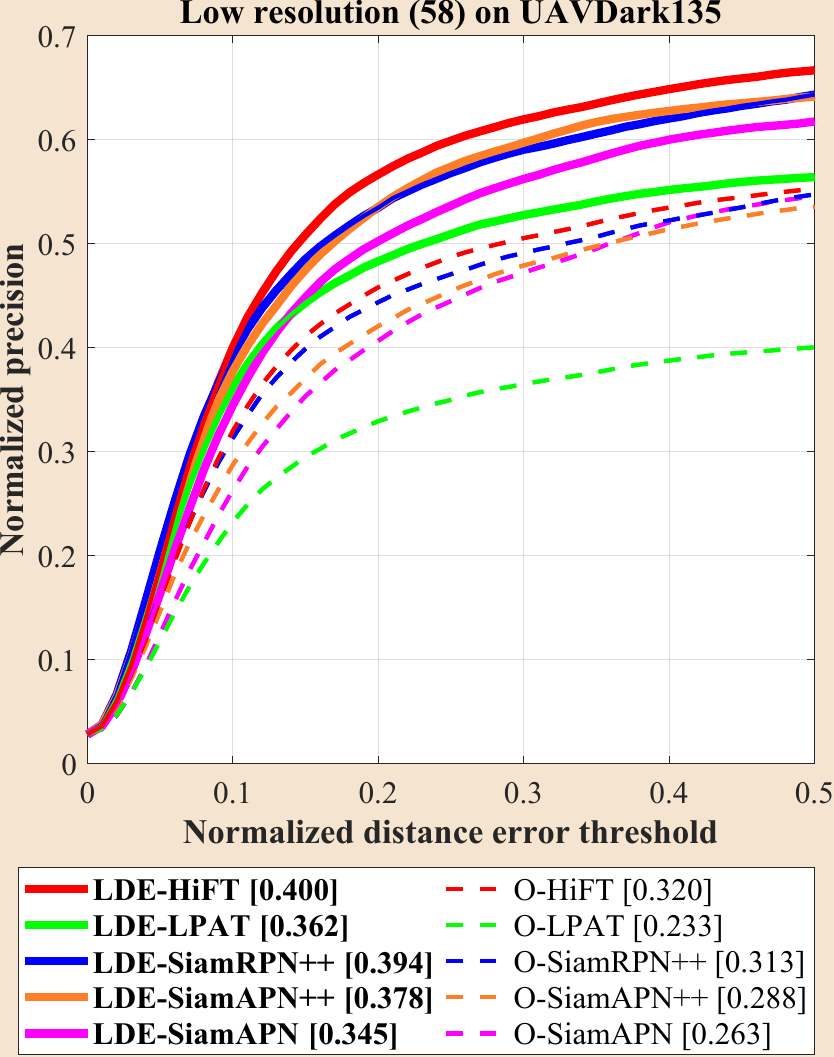}
	\end{minipage}
	\begin{minipage}{0.235\linewidth}
		\centering
		\includegraphics[width=1.0\linewidth]{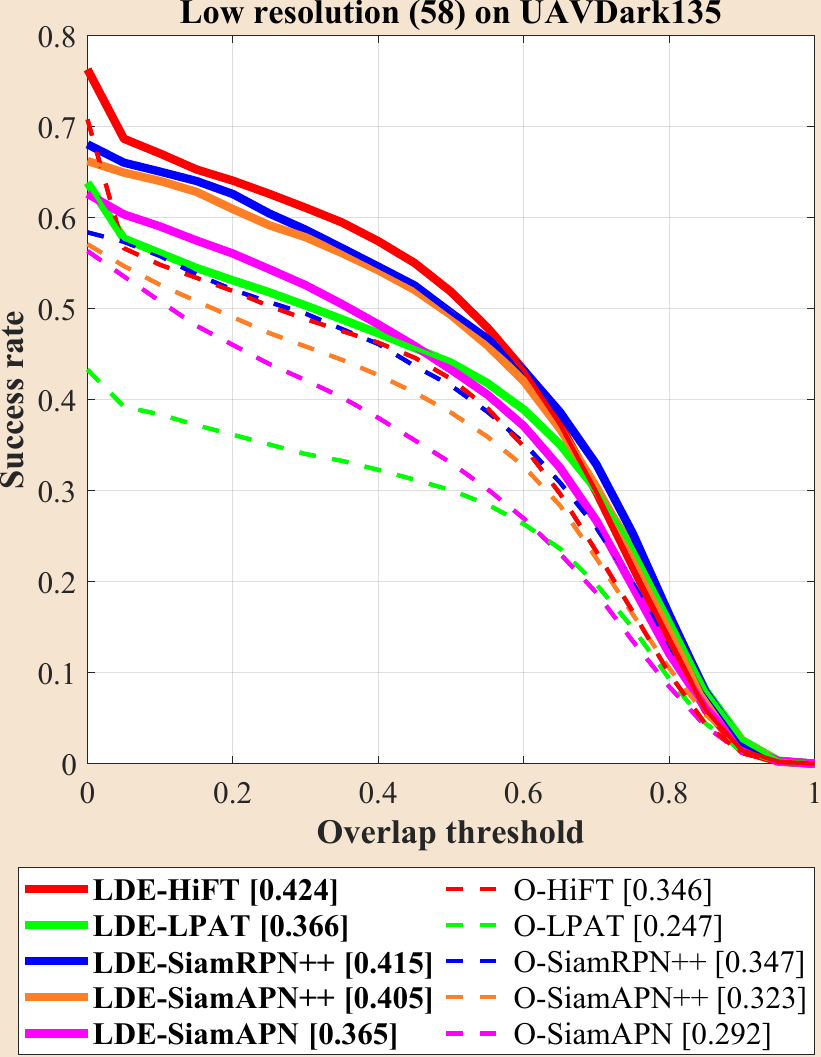}
	\end{minipage}}
    \caption{Normalized precision plots and success plots of illumination variation and low-resolution attributes of SOTA UAV trackers with LDEnhancer utilized (plots with solid lines) or not (plots with dashed lines) on UAVDark135.}
	\label{fig:attr}
 \vspace{-10pt}
\end{figure*}

\subsection{Evaluation Metrics}
The proposed enhancer serves for nighttime UAV tracking. Evaluation metrics in visual object tracking are applied to evaluate the enhancer. The three metrics in the one-pass evaluation (OPE) metrics~\cite{muller2018trackingnet} are precision, normalized precision, and success rate. Precision is determined by the center location error (CLE), and the fraction of frames with a CLE lower than 20 pixels is displayed as the precision plot (PP). The success rate is measured using the intersection over union (IoU), and the percentage of frames with an IoU greater than the configured maximum threshold is shown as the success plot (SP). The area under the curve (AUC) of the success plot is used to rank the success rate of trackers. To account for variations in image resolution and bounding box scale, the normalized precision metric is selected.

\subsection{Effectiveness and Superiority of LDEnhancer for Nighttime UAV Tracking}
\textit{1)\quad LDEnhancer with Different Trackers:}
To prove the effectiveness and superiority of the proposed method, the enhancer is tested on the authoritative night tracking datasets, \textit{i.e.}, UAVDark135, and the proposed  NAT2024-2, combined with 5 typical SOTA trackers, including HiFT~\cite{cao2021hift}, SiamRPN++~\cite{li2019siamRPN++}, SiamAPN++~\cite{cao2021siamapn++}, SiamAPN~\cite{fu2021onboard}, and LPAT~\cite{peng2022lpat}.
The objects in UAVDark135 undergo various light conditions, making the uneven light distribution problem more serious. As illustrated in Fig.~\ref{fig:all}, the proposed method further strengthens the adaptability of the SOTA UAV trackers in the nighttime. The enhancer has significantly improved the performance of SiamAPN++ for nighttime UAV tracking: \textbf{25.0\%} on precision, \textbf{29.8\%} on normalized precision, and \textbf{25.5\%} on success rate. Moreover, despite the adverse impact of complex nighttime conditions on the performance of LPAT, the implementation of LDEnhancer enables this tracker to achieve superior outcomes.


Evaluations visualization: as shown in Fig.~\ref{fig:line}, the presented screenshots visually demonstrate the effectiveness of the proposed method. The results indicate that the application of
LDEnhancer significantly enhances the performance of the SOTA trackers during low-light conditions.

\textit{2)\quad Comparison with SOTA Low-Light Enhancers:}
The comparison with other SOTA enhancement methods is shown in TABLE \ref{tab:4}. To present the advantage of LDEnhancer for nighttime UAV tracking, it is compared with other nine SOTA low-light enhancers utilizing the same tracker SiamRPN++ on the UAVDark135 benchmark, including RUAS~\cite{liu2021retinex}, LIME~\cite{guo2016lime}, EnlightenGAN~\cite{jiang2021enlightengan}, LLVE~\cite{zhang2021learning}, DCE++~\cite{li2021learning}, KinD++~\cite{zhang2021beyond}, DarkLighter~\cite{ye2021darklighter}, HighlightNet~\cite{fu2022highlightnet}, and SCT~\cite{ye2022tracker}. The results demonstrate the impressive performance of LDEnhancer in nighttime UAV tracking, which is attributed to its light distribution suppression. Conversely, other general low-light enhancers exhibit suboptimal performance. Notably, LDEnhancer raises the SiamRPN++ by \textbf{16.7\%} and \textbf{18.1\%} in success rate and precision, respectively, surpassing the second-best performer, SCT. 

\Remark Although these enhancers, \textit{i.e.}, DarkLighter, HighlightNet, and SCT, are also nighttime UAV tracking-oriented, the inferior performance exposes the weakness of common enhancers in the face of uneven light distribution.

\begin{table}[!b]
    \vspace{-10pt}
    \centering
    \caption{Comparison of LDEnhancer with SOTA enhancers on UAVDark135~\cite{li2022all}. The first results are in \textbf{bold}.}	 \setlength{\tabcolsep}{3mm}{
    \colorbox{table_c}{
    \resizebox{0.46\textwidth}{!}{
    \begin{tabular}{ccccc}
         \toprule
         UAVDark135 & Succ. &  Prec.  & $\Delta_{s} (\%)$ & $\Delta_{p} (\%)$ \\
         \midrule
         SiamRPN++ & 0.372 & 0.474 & - & -\\
         \midrule
         +RUAS~\cite{liu2021retinex} & 0.381 & 0.490 & 2.419 & 3.376 \\
         +KinD++~\cite{zhang2021beyond} & 0.386 & 0.485 & 3.763 & 2.321 \\
         +LIME~\cite{guo2016lime} & 0.393 & 0.501& 5.645 & 5.696 \\
         +EnlightenGAN~\cite{jiang2021enlightengan}& 0.397 & 0.505 & 6.720 & 6.540 \\
         +LLVE~\cite{zhang2021learning}& 0.397 & 0.508 & 6.720 & 7.173 \\
         +DCE++~\cite{li2021learning}& 0.402 & 0.512 & 8.065 &8.017 \\
        
        \midrule
        +Darklighter~\cite{ye2021darklighter} & 0.385 &0.495 & 3.495 & 4.430
\\
        +HighLightNet~\cite{fu2022highlightnet} & 0.376 & 0.483 & 1.075 & 1.899 
\\
        +SCT~\cite{ye2022tracker} & 0.421 & 0.547 & 13.172 & 15.401 \\
        \midrule

        \textbf{+LDEnhancer(Ours)} &\textbf{0.434} & \textbf{0.560} & \textbf{16.667} & \textbf{18.143} 
\\
         \bottomrule
    \end{tabular}
    }}}
    \label{tab:4}%
\end{table}
\begin{table}[b]
  \centering
  \vspace{-10pt}
  \caption{Ablation study of the LDEnhancer on UAVDark135. $\Delta$ reflects the improvement compared to the base tracker, \textit{i.e.}, SiamRPN++, with the highest performance highlighted in \textbf{bold}.
  }
  \vspace{-5pt}
  \renewcommand{\arraystretch}{1.2} 
  \resizebox{\linewidth}{!}{
  \colorbox{table_c}{
    \begin{tabular}{lcccc}
    \hline
    Trackers & Succ. & $\Delta_{s} (\%)$ & Prec. & $\Delta_{p} (\%)$\\
    \hline
    SiamRPN++ & 0.372 & - & 0.474 & -  \\
    +Baseline & 0.395 & +6.183 & 0.510 & +7.595  \\
    
    +Baseline+TB & 0.408 & +9.677 & 0.529 & +11.603 \\
    +Baseline+TB+LS& 0.421 & +13.172 & 0.541 & +14.135 \\
    \hline
    \textbf{+Baseline+TB+LS+TF(LDEnhancer)}  & \textbf{0.434} & \textbf{+16.667} & \textbf{0.560} & \textbf{+18.143} \\
    \hline
    \end{tabular}
    }}
  \label{tab:ab}%
\end{table}
\textit{3)\quad Attribute-Based Performance:}
Additional environmental changes caused by illumination variation and low resolution can exacerbate the difficulty of nighttime UAV tracking. To thoroughly assess the effectiveness of LDEnhancer, a comparison of their pertinent properties is conducted. Figure~\ref{fig:attr} proves the robustness of our framework in several challenging conditions. SiamAPN++ raises the success rate of the existing best performance by \textbf{25.7\%} on UAVDark135 for illumination variation. In addition, SiamAPN++ realizes a success rate of \textbf{0.405} for low resolution on UAVDark135, which improves the existing best performance by \textbf{25.4\%} for nighttime UAV tracking.

\subsection{Ablation Study}
Different variants of LDEnhancer are evaluated with the same UAV tracker, \textit{i.e.}, SiamRPN++, on UAVDark135 in this subsection. Specifically, utilizing one branch structure to estimate parameter maps is denoted as the Baseline. TB stands for utilizing two branches structure to estimate different parameter maps. LS represents employing the unsupervised light distribution loss. TF denotes employing the cross-attention. As shown in TABLE~\ref{tab:ab}, with enhancers, the performance of UAV trackers is significantly improved. 
\begin{figure}[!t]
    \vspace{6pt}
	\centering	
		\includegraphics[width=1.0\linewidth]{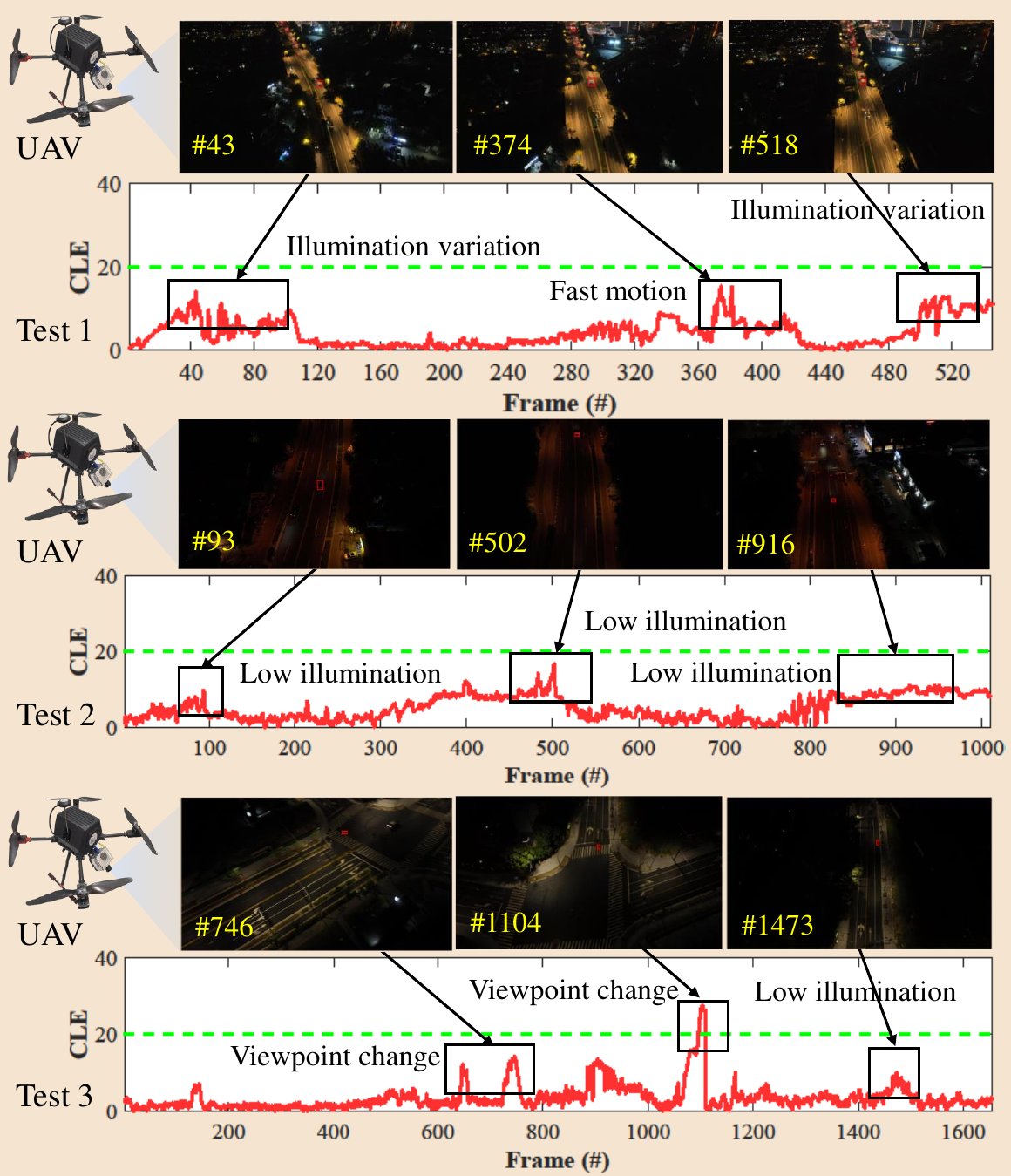}
	\caption
	{Several results of real-world tests. The estimated positions are represented by the \textcolor{red}{red} bounding boxes. The CLE below the \textcolor[rgb]{0,1,0}{green} dashed line is the success tracking in the real-world test. With LDEnhancer, the tracker can effectively realize robust nighttime UAV tracking.
	}
	\label{fig:real}
\vspace{-10pt}
\end{figure}

 \subsection{Real-World Tests} 
To demonstrate the applicability of LDEnhancer in real-world UAV tracking, extensive real-world tests are adopted on a typical UAV platform equipped with an NVIDIA Orin NX-based edge camera. The SiamRPN++ enabled with LDEnhancer exceeding \textbf{30.3} frames per second. As shown in Fig.~\ref{fig:real}, the main challenges in the tests are illumination variation, low illumination, fast motion, and viewpoint change. In Test 1, the primary challenge is the truck moving in the bright area and the shadow area with a frequent illumination variation. In Test 2, the poor light condition exacerbates the difficulties associated with robust tracking. Test 3 presents viewpoint change and low illumination. The CLE curves show that the prediction errors are within 20 pixels, which can be regarded as reliable nighttime tracking. 

\section{Conclusions}
This work proposes a novel enhancer, LDEnhancer, enhancing nighttime UAV tracking with light distribution suppression. Specifically, a novel image content refinement module and a new light distribution generation module are proposed to capture light distribution information and image content information effectively. Two different parameter estimation modules are constructed to process the features with different information, respectively, for the parameter map prediction. Leveraging two parameter maps, an innovative interweave iteration adjustment is proposed for the collaborative pixel-wise adjustment of low-light images. Experimental results prove the effectiveness as well as the efficiency of our enhancer. We believe this work would contribute to the advancement of nighttime UAV tracking.
%
%

\section*{Acknowledgment}

This work is supported by the National Natural Science Foundation of China (No. 62173249) and the Natural Science
Foundation of Shanghai (No. 20ZR1460100).

\bibliographystyle{IEEEtran}
\normalem
\balance
\bibliography{IROS2024}
\end{document}